\documentclass[letterpaper, 10 pt, conference]{ieeeconf}  

\IEEEoverridecommandlockouts                              

\usepackage{tabularx}
\usepackage{graphicx}
\usepackage{hyperref} 
\usepackage{flushend}
\usepackage{amsmath}
\usepackage{amsfonts}
\usepackage{array,booktabs}

\usepackage{rotating, multirow}
\usepackage{makecell}

\hypersetup{
    colorlinks=true,
    linkcolor=black,
    filecolor=black,      
    urlcolor=black
    }

\title{\LARGE \bf
Improved Benthic Classification using Resolution Scaling and \\SymmNet Unsupervised Domain Adaptation}

\author{Heather Doig$^{1}$, Oscar Pizarro$^{1,2}$ and Stefan B. Williams$^{1}$
\thanks{$^{1}$The authors are with the Australian Centre for Field Robotics, University of Sydney, NSW Australia, {\tt\small(h.doig, o.pizarro, stefanw)@acfr.usyd.edu.au}}
\thanks{$^{2}$O. Pizarro is also with the Marine Technology Department, Norwegian University of Science and Technology, Trondheim, Norway.
}
}

\begin{document}

\maketitle
\thispagestyle{empty}
\pagestyle{empty}

\begin{abstract}
Autonomous Underwater Vehicles (AUVs) conduct regular visual surveys of marine environments to characterise and monitor the composition and diversity of the benthos.  The use of machine learning classifiers for this task is limited by the low numbers of annotations available and the many fine-grained classes involved.  In addition to these challenges, there are domain shifts between image sets acquired during different AUV surveys due to changes in camera systems, imaging altitude, illumination and water column properties leading to a drop in classification performance for images from a different survey where some or all these elements may have changed.  This paper proposes a framework to improve the performance of a benthic morphospecies classifier when used to classify images from a different survey compared to the training data.  We adapt the SymmNet state-of-the-art Unsupervised Domain Adaptation method with an efficient bilinear pooling layer and image scaling to normalise spatial resolution, and show improved classification accuracy.  We test our approach on two datasets with images from AUV surveys with different imaging payloads and locations.  The results show that generic domain adaptation can be enhanced to produce a significant increase in accuracy for images from an AUV survey that differs from the training images.  
\end{abstract}

\section{Introduction}
\label{sec:intro}

Autonomous Underwater Vehicles (AUVs) are used to conduct regular visual surveys of the marine environment to measure and monitor changes in the benthic environment due to stresses such as pollution, over-fishing and climate change~\cite{Monk2018Evaluation,Perkins2021AnalysisOfATime}. Typically, marine scientists label the images with point annotations to measure the presence and diversity of benthic species and physical features using a morphological hierarchy such as in \cite{Pavoni2019Challenges}.  Machine learning classifiers promise to make this process more efficient and increase the amount of information gained from a survey \cite{GonzalezRivero2020Monitoring, Williams2019Leveraging} but there are typically only small amounts of labelled annotations available for training.

Classifiers of morphospecies do not transfer well between image datasets due to domain shift between different AUV surveys \cite{ Pavoni2019Challenges, Langenkamper2020,Williams2019Leveraging}.  Domain shift arises from differences in cameras, imaging altitude, illumination and water column properties. In addition, classifying benthic species and physical features is a fine-grained classification problem as there are many classes with high intra-class variations and also similar inter-class features, increasing the difficulty of the classification task \cite{AniBrownMary2017CoralReefImage,Zhu2022DualCross}.

    \begin{figure}[!t]
        \centering
        \includegraphics[width=0.47\textwidth]{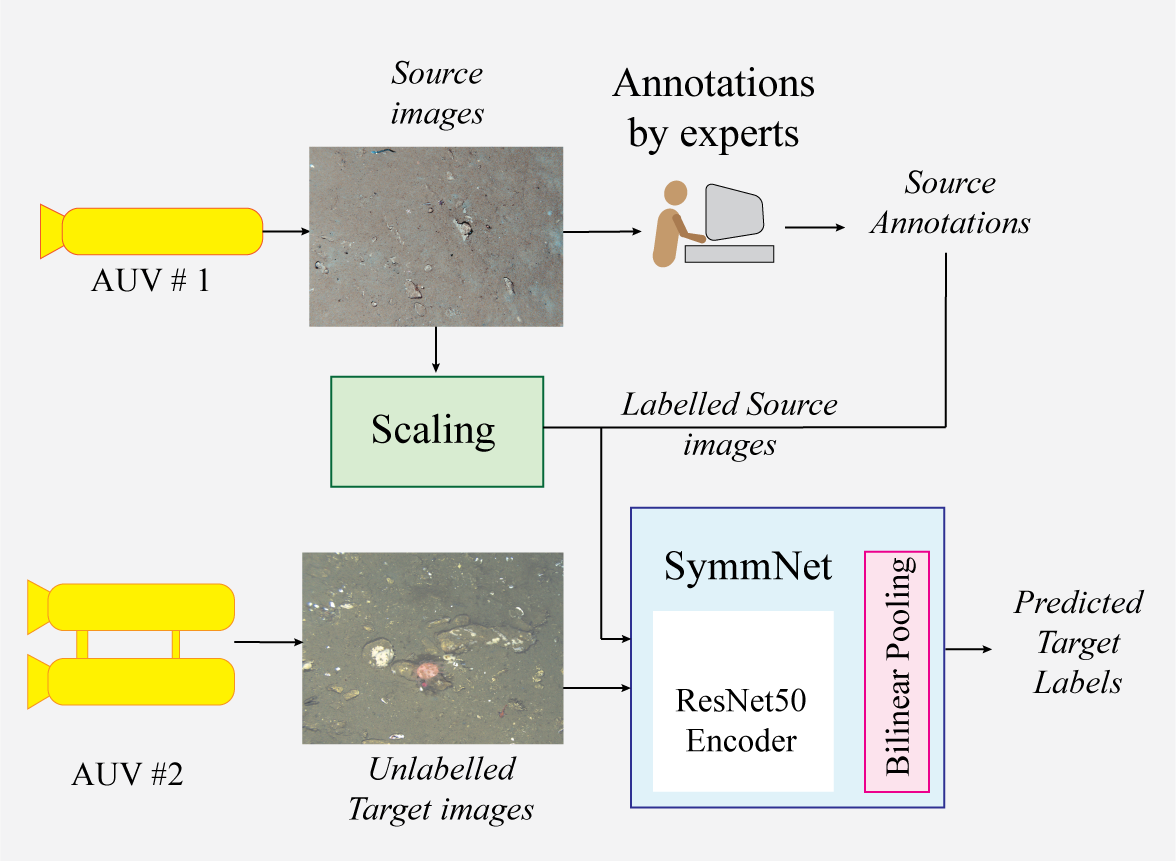}
        \caption{\textbf{Overview}  Framework to improve the transferability of a benthic morphospecies classifier when using labelled source images from one AUV survey for training and testing on unlabelled target images from another AUV survey.  The framework includes a) scaling images to match spatial resolution between source and target b) using Unsupervised Domain Adaptation (SymmNet, \cite{Zhang2019Domain}) on the labelled source images and unlabelled target images to train a classifier for source and target images and c) using an efficient bilinear pooling layer \cite{Yu2022Efficient} for more discriminative features.  The framework improves classification accuracy compared to training with the unscaled source images alone.}
        \label{pipeline}
    \end{figure}
    
To address the challenges of making more efficient use of AUV imagery with scarce label annotations, this paper proposes a framework to explore and understand the use of Unsupervised Domain Adaptation (UDA)~\cite{Gong2014Learning,Ganin2015DomainAdversarial} with the aim of increasing performance of classifiers trained with data from different AUV surveys.  This process would allow labelling effort for images from one AUV survey to be migrated to a new survey with new operating parameters arising from changes to cameras, lighting or imaging altitude.  Classification performance will also be improved by scaling images so that images have the same spatial resolution \cite{Walker2019EffectOfPhysics} and adding an efficient bilinear pooling layer to generate more discriminative features \cite{Yu2022Efficient}.  Figure \ref{pipeline} illustrates the proposed framework.

Performance of a machine learning model trained on a source domain, such as a classifier, can drop when applied to a target domain due to a shift in the distribution of the features.  UDA reduces the difference between labelled source domain data and unlabelled target domain data.  A successful range of methods update the feature representation for the target domain using either statistics of the distributions \cite{Gong2014Learning,Long2017DeepTransferJAN} or adversarial training \cite{Ganin2015DomainAdversarial,Tzeng2017AdvDiscDomainAdapt}.

Adversarial UDA uses a domain discriminator to distinguish between features from the source and target domain.  An adversarial loss updates the network during training to align the features from the source and target domain.  This paper uses state-of-the-art UDA, SymmNet \cite{Zhang2019Domain}, which has demonstrated success on a similarly challenging problem of classifying aerial habitats from different drone imagery \cite{Nagananda2021Benchmarking}.  SymmNet uses an adversarial loss applied at the class level producing alignment of the source and target distribution at both the domain and class level.


This paper makes the following contributions:
\begin{itemize}
    \item We propose a framework using SymmNet UDA, and scaling to improve the classification of point annotations from images taken from different AUV platforms or locations.
    \item We explore the impact to classification performance when an efficient bilinear pooling layer that requires a small number of parameters with fast inference time is applied.
    \item We provide results and analysis from two datasets from benthic surveys both with images from two different AUV surveys.
    \item We introduce two curated datasets of benthic image patches to support further research into UDA between AUV surveys\footnote{\url{https://data.mendeley.com/datasets/d2yn52n9c9}}.
\end{itemize}

The remainder of the paper is structured as follows. Section \ref{sec:related} describes related work in Unsupervised Domain Adaptation, bilinear pooling and scaling spatial resolution.  Section \ref{sec:method} describes the implementation of SymmNet and Two-Level Kronecker Product Factorization bilinear pooling as well as the two datasets used to demonstrate the framework.  The results of the experiments are presented in Section \ref{sec:result} followed by some concluding remarks and suggestions for future research in Section \ref{sec:conclusion}.  Our code is available at {\url{https://github.com/hdoi5324/benthic-uda}}.

\section{Related Work}
\label{sec:related}

\subsection{Unsupervised Domain Adaptation}
UDA is a machine learning technique used to reduce the domain shift between labelled \emph{source} data and unlabelled \emph{target} data, such that a domain-adapted classifier originally trained with labelled source data can be used with the target data.  The difference in the distribution between source and target data can been reduced by aligning the feature representations using higher order statistics \cite{Gong2014Learning,Long2017DeepTransferJAN} or adversarial training \cite{Ganin2015DomainAdversarial,Tzeng2017AdvDiscDomainAdapt}.  We focus on adversarial training techniques which have delivered state-of-the-art results for image classification. \cite{Ganin2015DomainAdversarial} and \cite{Tzeng2017AdvDiscDomainAdapt} align the feature representations using a discriminator network to distinguish samples between domains while using a confusion loss to adapt the target encoder.  In this paper, we use SymmNet \cite{Zhang2019Domain,Zhang2020UMulti} which replaces the domain discriminator with a multi-class discriminator loss giving more separable features at a class level.

\subsection{Bilinear Pooling}
Bilinear pooling has been shown to provide more discriminative features from a deep learning Convolutional Neural Network (CNN) model \cite{Lin2015BilinearCNN,Kong2017LowRank}.  Applying this pooling layer between CNN modules or after the final CNN module has increased the accuracy of fine-grained and texture based classification tasks \cite{Liu2019FromBow}.  For example, \cite{Wang2020Attention} improved the retrieval of remote sensing images by adding a bilinear pooling layer after the CNN backbone.  The resulting features provided more detail about the scene despite variations in atmospheric conditions, illumination and viewing angles which is similar to the variations that arise in benthic images.  This work uses Two-Level Kronecker Product Factorization bilinear pooling \cite{Yu2022Efficient} which aims to use fewer parameters than other implementations \cite{Lin2015BilinearCNN,Wang2020Attention}.  Fewer parameters reduces computational resources and inference time which may allow the method to be used onboard an AUV providing real-time results.


\subsection{Scaling}
Adjusting images to a common spatial resolution can increase the performance of a CNN when a small number of annotations are available for training \cite{Zurowietz2020UnsupervisedKnowledge,Walker2019EffectOfPhysics}.  Spatial resolution may differ between AUV surveys due to changes to the camera (e.g. imager size and/or resolution, lens field of view, etc)  and altitude used during the mission.  \cite{Walker2019EffectOfPhysics} performed scale adjustment on images from different AUVs using altitude and spatial resolution per pixel, improving the generalization error in an image segmentation task.

\section{Method}
\label{sec:method}
    \begin{figure}[t!]
        \centering
        \includegraphics[width=0.47\textwidth]{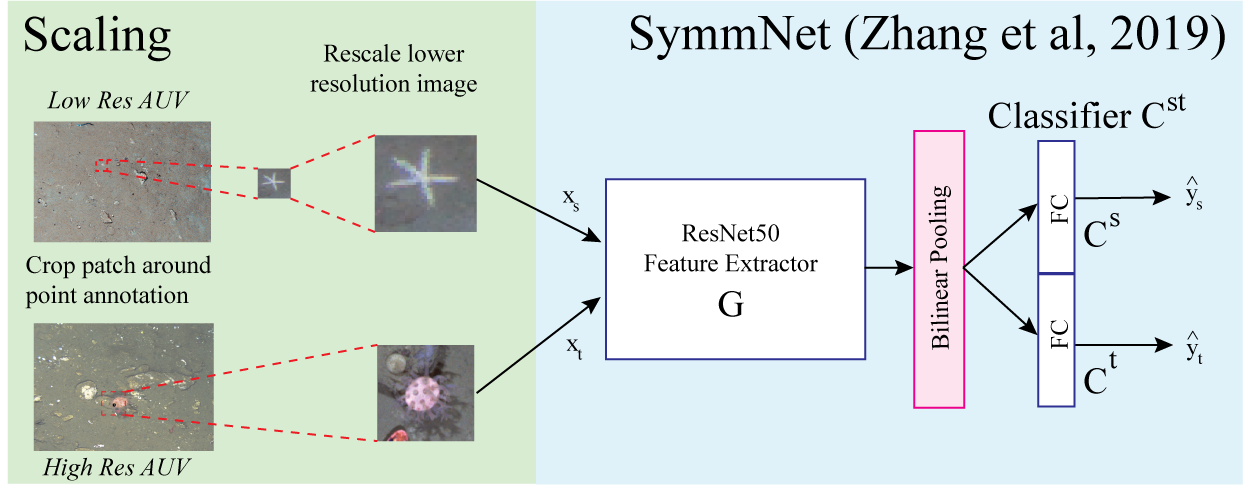}
        \caption{Network diagram for full framework.  Scaling of the lower resolution AUV patch is performed prior to training.  The SymmNet network uses ResNet50 backbone for the encoder followed by the bilinear pooling layer then the symmetric source and target classifier.  The bilinear pooling layer uses a Two-Level Product Factorization to calculate a projection of the bilinear product of the feature map from the encoder.}
        \label{network}
    \end{figure}

The proposed framework is comprised of a classifier network based on SymmNet \cite{Zhang2019Domain}.  The classifier has a ResNet50 backbone \cite{He2016Deep} with the pooling layer after Layer 4 CNN module replaced with a bilinear pooling layer described below. Figure \ref{network} provides an overview of the three main components of the framework being scaling, the SymmNet classifier and bilinear pooling which will now be described in more detail.

\subsection{Scaling}
The input to the network is an image patch cropped around a human-generated point annotation and scaled to the same spatial resolution and pixel size (224x224) for the source and target images.  Table \ref{tab:auv_data} provides the spatial resolutions for the original images and the crop size used.

\subsection{SymmNet Classifier}
\label{subsec:symmnet}
The Domain-Symmetric Network or SymmNet by \cite{Zhang2019Domain} is an UDA network that uses adversarial training to adapt a target classifier using labelled source data and unlabelled target data.   The network uses a symmetric design with a combination of loss terms using domain and class level confusion loss to improve the adaptation of the target classifier to the target domain distribution as well as the class distribution, improving on previous methods such as \cite{Tzeng2017AdvDiscDomainAdapt}.  

The network comprises a shared feature extractor $G$ using a ResNet50 backbone and pooling layer \cite{He2016Deep} followed by a fully connected classifier $C^{st}$ that provides classifications for both target and source images.   Its symmetric design combines a source classifier $C^s$ and the target classifier $C^t$ each with $K$ outputs where $K$ is the number of classes.  When using bilinear pooling, the average pooling layer after Layer 4 of ResNet is replaced with the bilinear pooling described in the next section.  

The following detail provides a brief overview of the losses that are used to train the feature extractor and classifier while also adapting the target classifier to the unlabelled target data.  $x$ denotes the input data being the image and $y$ is the label for the input image.  The source data is defined as $D_s={(\mathbf{x}_i^s,y_i^s)}^{n_s}_{i=1}$ with $n_s$ labelled samples and the unlabelled target data is $D_{t}={(\mathbf{x}_j^t)}^{n_t}_{j=1}$ with $n_t$ unlabelled samples. 

Firstly, the classifiers $C^s$ and $C^t$ are each trained using the labelled source data with a cross-entropy loss as is common for a classifier:
\begin{equation}
\min_{C^{dom}} E^{dom}_{cls} (G,C^{dom}) = -\frac{1}{n_s} \sum^{n_s}_{i=1} log(p^{dom}_{y^s_i}(\mathbf{x}^s_i))
\label{eq:Cls_loss}
\end{equation}
where $p^{dom} = C^{dom}(G(\mathbf{x}))$ and $dom$ is the source $s$ or target $t$ domain.
Domain discrimination in the next loss terms adapt the target classifier $C^t$ so that it differs from the source classifier $C^s$.

Instead of using a separate domain discriminator network as in \cite{Tzeng2017AdvDiscDomainAdapt}, the discrimination training uses the sum of probabilities for source samples through $C^s$ and target samples through $C^t$ to train the $C^{st}$ classifier using the following two-way cross-entropy loss:
\begin{multline}
    \min_{C^{st}} E^{st}_{dom} (G,C^{st}) = -\frac{1}{n_t} \sum^{n_t}_{j=1} log(  \sum^{K}_{k=1} p^{st}_{k+K}(\mathbf{x}^t_j)) \\ - \frac{1}{n_s} \sum^{n_s}_{i=1} log(  \sum^{K}_{k=1} p^{st}_{k}(\mathbf{x}^s_i))
\label{eq:Domain_st_loss}
\end{multline}
where $p^{st} = C^{st}(G(\mathbf{x}))$ and $K$ is the number of classes.  The source and target classifiers now discriminate between source and target samples.

The following confusion loss term discriminates at the class level between the source and target domain leading to better adaptation of the target classifier at a class level.
\begin{multline}
    \min_{G} F^{st}_{class} (G,C^{st}) = -\frac{1}{2n_s} \sum^{n_s}_{i=1} log(p^{st}_{y^s_i+K}(\mathbf{x}^s_i)) \\ -\frac{1}{2n_s} \sum^{n_s}_{i=1} log(p^{st}_{y^s_i}(\mathbf{x}^s_i)) 
\label{eq:G_st_loss}
\end{multline}

A domain level confusion loss using the unlabelled target data compares the sum of probabilities from source and target classifiers of the combined $C^{st}$ classifier using the target data: 
\begin{multline}
    \min_{G} F^{st}_{dom} (G,C^{st}) = -\frac{1}{2n_t} \sum^{n_t}_{j=1} log(  \sum^{K}_{k=1} p^{st}_{k+K}(\mathbf{x}^t_j)) \\ - \frac{1}{2n_t} \sum^{n_t}_{j=1} log(\sum^{K}_{k=1} p^{st}_{k}(\mathbf{x}^t_j))
\label{eq:dom_t_loss}
\end{multline}

The final loss uses an entropy minimization objective to increase the classification task performance by summing the probabilities for each matching class in the combined $C^{st}$ classifier:
\begin{equation}
\min_{G} M^{st}(G, C^{st})=-\frac{1}{n_t}\sum^{n_t}_{j=1} \sum^{K}_{k=1} q^{st}_k(\mathbf{x}^t_j)log(q^{st}_k(\mathbf{x}^t_j))
\end{equation}
\label{eq:entropy_loss}
where $q^{st}_k(\mathbf{x}^t_j)=p^{st}_k(\mathbf{x}^t_j)+p^{st}_{K+k}(\mathbf{x}^t_j),k\in\{1,\cdots,K\}$.  

These losses combine to create two overall losses to minimise:
\begin{multline}
 \min_{C^s, C^t, C^{st}} E^s_{cls} (G,C^s) + E^t_{cls} (G,C^t) + E^{st}_{dom} (G,C^{st}),\\
\min_{G} F^{st}_{class} (G,C^{st}) + \lambda(F^{st}_{dom} (G,C^{st}) + M^{st}(G, C^{st}))
\end{multline} 
\label{eq:total_G_loss}
where $\lambda\in[0,1]$ is a trade-off parameter that is increased during training.

\subsection{Bilinear Pooling}
\label{subsec:bp}

\begin{figure}[t!]
        \centering
        \includegraphics[width=0.47\textwidth]{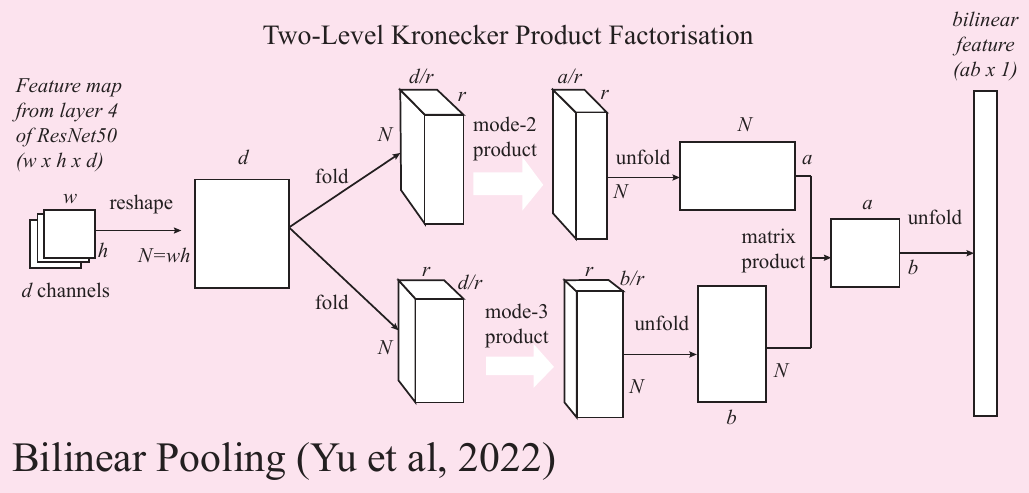}
        \caption{Network diagram for the bilinear pooling layer with Two-level Kronecker Product Factorization.}
        \label{network_bp}
    \end{figure}

Bilinear pooling calculates a more discriminative feature $\mathbf{b}$ from the outer product of an input feature map $(\mathbf{XX^T})$ but this approach is highly memory intensive.  The bilinear pooling layer in this paper uses the low parameter count Two-Level Kronecker Product Factorization (TKPF) method \cite{Yu2022Efficient} to approximate a projection of the bilinear product.  The configuration used in this work adds only 4K parameters to the ResNet50 backbone which has 23M parameters.  Two smaller scale matrices $\mathbf{A}$ and $\mathbf{B}$ are learned to create a projection of the bilinear product of the input feature map $\mathbf{X}$.   The network is repeated $q$ times and the results are averaged to increase the representative capability. The parameters  $a$, $b$, $r$ and $q$ are set at training.  This process is visualized in Figure \ref{network_bp} and a high-level description is provided below.  The complete factorization method can be found in \cite{Yu2022Efficient}.  

A projection matrix $\mathbf{P}$ can be used to generate a lower dimension bilinear feature:
\begin{equation}
    \mathbf{b} = \mathbf{P}vec(\mathbf{XX}^T)
\label{eq:b_projection}
\end{equation}
where $\mathbf{X}$ is the feature map from the last layer of ResNet. $\mathbf{P}$ is constrained to be a Kronecker product as this reduces a very larger matrix to two smaller matrices:
\begin{equation}
   \mathbf{P} = \mathbf{A} \otimes \mathbf{B} 
\end{equation}
By substituting in the Kronecker product into Equation \ref{eq:b_projection}, there can be a further factorization using the `vec' trick described in \cite{Yu2022Efficient}:
\begin{equation}
    (\mathbf{A} \otimes \mathbf{B})vec(\mathbf{XX}^T)= vec(\mathbf{ST}^T)
\end{equation}
where $\mathbf{S}=\mathbf{BX}$ and  $\mathbf{T}=\mathbf{AX}$.  A second level of Kronecker product can be used to factorize further:
\begin{equation}
    \mathbf{A} = \mathbf{I_r} \otimes \hat{\mathbf{A}}, \mathbf{B} = \hat{\mathbf{B}} \otimes \mathbf{I_r}
\end{equation}
allowing $\mathbf{S}$ and $\mathbf{T}$ to be calculated efficiently using modal folding and tensor modal products. $\mathbf{X}$ is folded into $\mathbf{\mathcal{X}}_a\in\mathbb{R}^{{{N}\times\frac{d}{r}}\times{r}}$ and $\mathbf{\mathcal{X}}_b\in\mathbb{R}^{{N}\times{r}\times\frac{d}{r}}$.  After folding, the tensors are multiplied using the tensor modal product:
\begin{align}
    \mathbf{T}  = \mathbf{\mathcal{X}}_a \times_2 \hat{\mathbf{A}} \\
    \mathbf{S} = \mathbf{\mathcal{X}}_b \times_3 \hat{\mathbf{B}}
\label{eq:modal}
\end{align}

The matrix product gives the projected bilinear feature:
\begin{equation}
    \hat{\mathbf{b}} = vec(\mathbf{S}\mathbf{T}^T)
\end{equation}
\label{eq:b_hat}
with $\hat{\mathbf{b}}\in\mathbb{R}^{ab}$.

Element-wise signed square root and L2 normalization is applied to $\hat{\mathbf{b}}$. The features from the $q$ duplicated networks are averaged giving the bilinear feature for the classifier. A dropout layer with probability of 0.7 is used before the classifier to add regularization.  Dropout of 0.7 is a common starting amount and was not tuned further.  

\subsection{Datasets}
The proposed framework was used on two datasets with images taken from two different AUV surveys~\cite{Doig2022}.

\subsubsection{Elizabeth and Middleton Reef}
\begin{figure}
\centering
\includegraphics[width=0.3\textwidth]{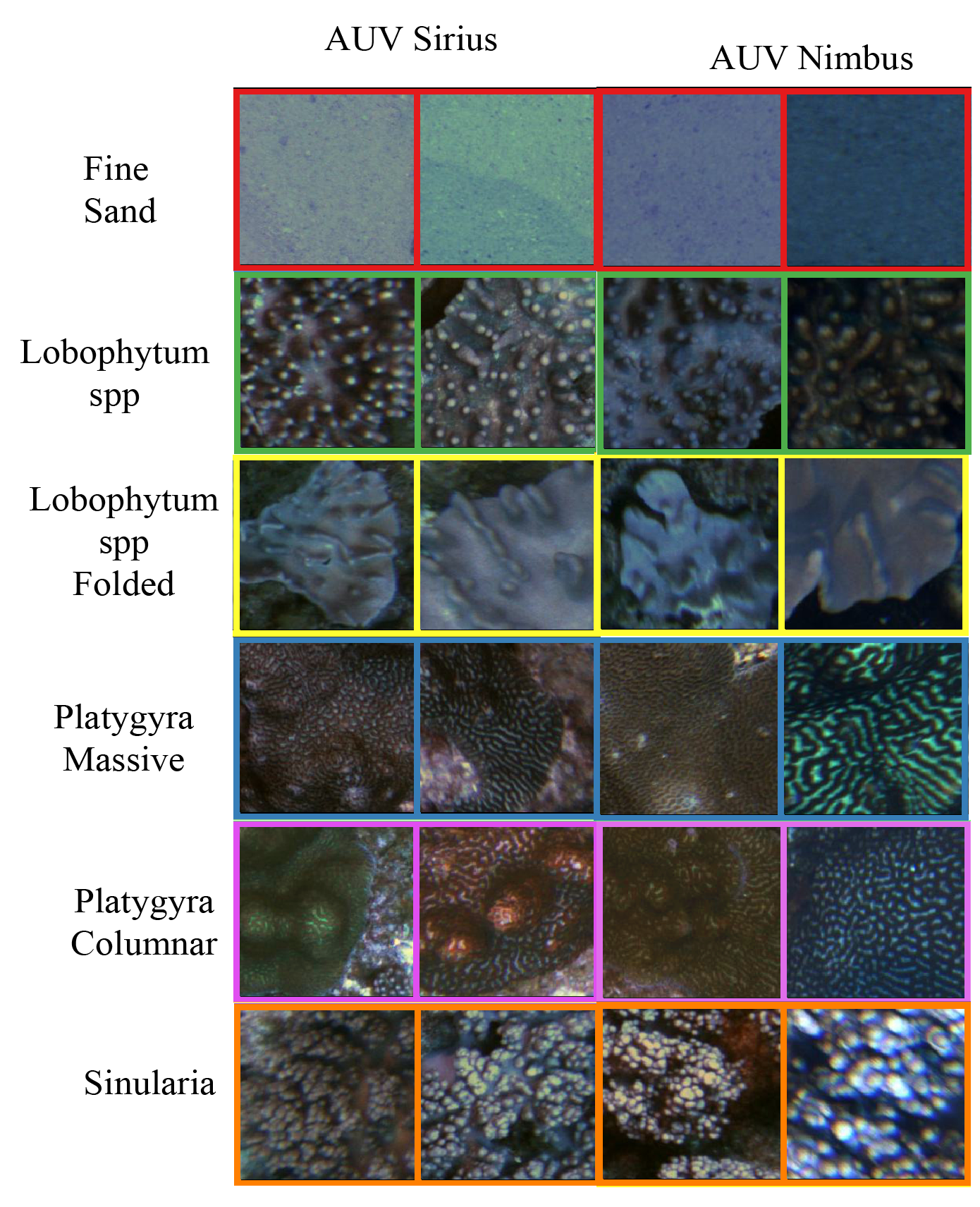}
\caption{Samples of the six classes from the EMR dataset. }
\label{emr_samples}
\end{figure}

The first dataset was collected at Elizabeth and Middleton Reef (EMR) during a marine environmental cruise completed in January 2020 using the AUV \emph{Sirius} and AUV \emph{Nimbus} \cite{Carroll2021ElizabethAndMiddleton}.  Manual point annotations of the five most common coral species as well as sand were used to create the dataset.  The images are from three separate surveys by the AUVs.  Figure \ref{emr_samples} shows examples of the six classes and Table \ref{tab:emr_data} shows the count of point annotations by class and survey.  

\begin{table}
\caption{Number of samples for EMR Dataset\label{tab:emr_data}}
\centering
\begin{tabular}{lccc}
\hline
Survey Name & NG06 & SS07 & SS09\\
\hline
AUV & \emph{Nimbus} & \emph{Sirius} & \emph{Sirius} \\
\hline
Fine Sand & 135 & 159 & 66 \\
Lobophytum spp & 153 & 199 & 60 \\
Lobophytum spp Folded & 152 & 194 & 70 \\
Platygyra Massive & 225 & 125 & 62 \\
Platygyra Columnar & 196 & 163 & 70 \\
Sinularia & 258 & 278 & 93 \\
\hline
Total Samples & 1119 & 1118 & 421 \\
\hline
\end{tabular}
\end{table}

\subsubsection{South Hydrate Ridge}
\begin{figure}
\centering
\includegraphics[width=0.3\textwidth]{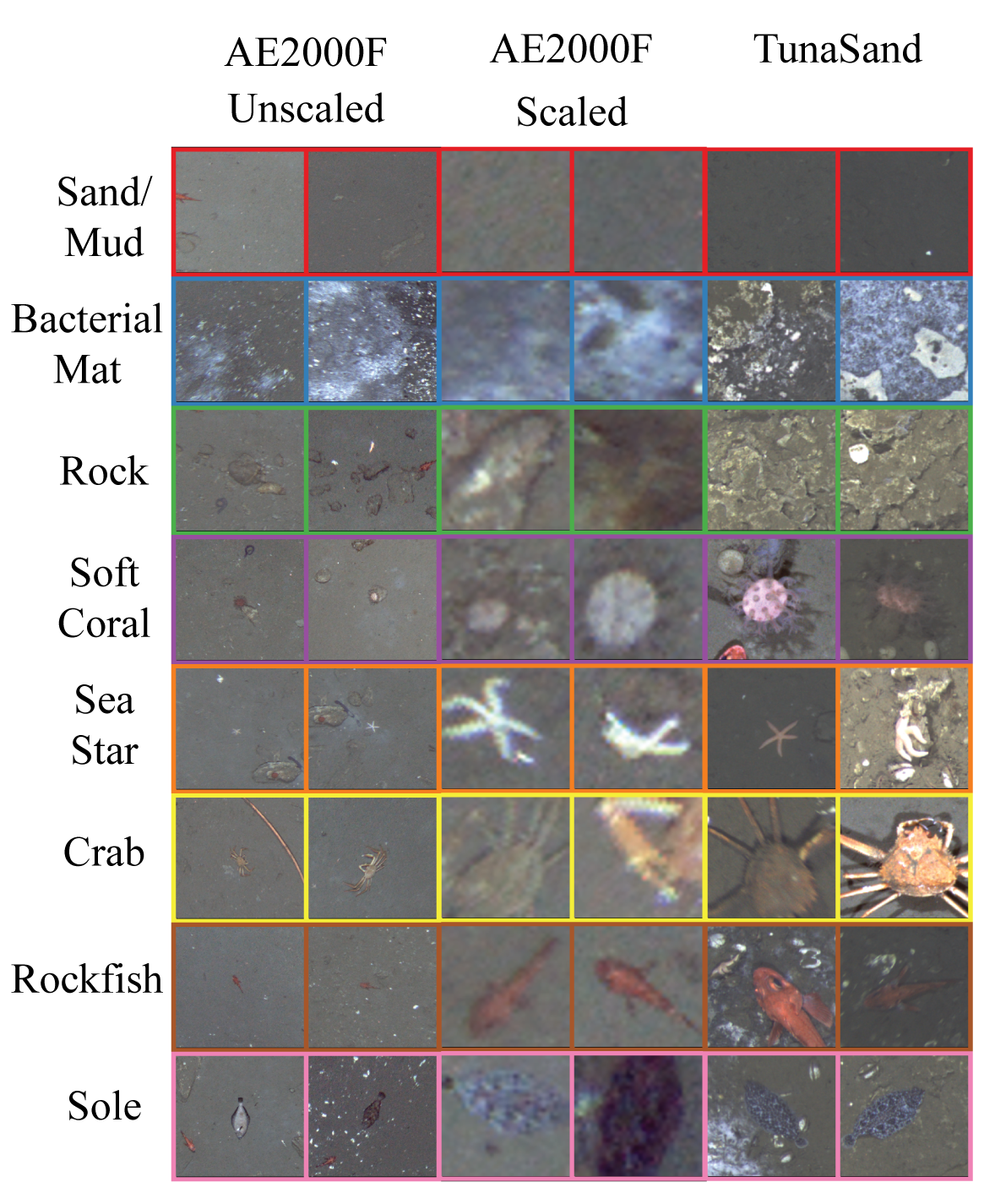}
\caption{Samples from the 8 classes in the SHR dataset.  Each row shows two unscaled patches from AUV \emph{AE2000f}, two patches from AUV \emph{AE2000f} scaled to the same spatial resolution as AUV \emph{Tuna-sand}, followed by two patches from AUV \emph{Tuna-sand}.  This shows how the same spatial scale displays species at similar size as seen in the Soft Coral and Rockfish samples.}
\label{shd_samples}
\end{figure}

The second dataset was from the South Hydrate Ridge (SHR) collected on the Adaptive Robotics cruise in the Eastern Pacific Ocean in 2018 \cite{Yamada2021LearningFeatures}. This dataset includes three physical features (Bacterial Mat, Sand/Mud and Rock) and five benthic species (Soft Coral, Sea Star, Crab, Rockfish and Sole) collected by the AUV \emph{AE2000f} and the AUV \emph{Tuna-sand} at depths of around 800 m.  The AUV \emph{AE2000f} has a very large image footprint operating at around 6m altitude while the AUV \emph{Tuna-sand} has high-resolution cameras working at 2 m altitude.  Figure \ref{shd_samples} provides samples of each class taken by the two AUVs as well as the scaled patch for the AUV \emph{AE2000f} and Table \ref{tab:shr_data} provides the number of samples in each class by AUV.  

\begin{table}
\caption{Number of samples for SHR Dataset
\label{tab:shr_data}}
\centering
\begin{tabular}{lcc}
\hline
AUV & \emph{AE2000f} & \emph{Tuna-sand}\\
\hline
Sand & 65 & 57 \\
Bacterial Mat & 60 & 60\\
Rock  & 62 & 54 \\
Soft Coral & 59 & 49 \\
Sea Star & 67 & 58 \\
Crab & 64 & 52 \\
Rockfish & 71 & 66 \\
Sole & 64 & 41 \\
\hline
Total Samples & 512  & 437 \\
\hline
\end{tabular}
\end{table}

\begin{table}
\caption{AUV Resolution\label{tab:auv_data}}
\centering
\begin{tabular}{lccc}
\hline
Dataset & AUV & Resolution (mm/pixel) & Scaled Crop (pixels) \\
\hline
EMR & Nimbus & 0.65 @ 2m & 148 \\
& Sirius & 0.43 @ 2m & 224 \\
\hline
SHR & AE2000f & 6 @ 6m & 32 \\
& Tuna-sand & 0.8 @ 2m & 224 \\
\hline
\end{tabular}
\end{table}

\subsection{Experiments}


For each dataset, each source and target pair was trained with all combinations of with or without scaling, bilinear pooling and SymmNet.  This resulted in eight possible combinations for the six source-target pairs.  When SymmNet was not used, only the losses in Equation \ref{eq:Cls_loss} and \ref{eq:G_st_loss} were used for training with the labelled source data.

No scaling used a crop of 224x224 pixels for both domain datasets.  When scaling was used, it was applied to the domain with the lower resolution by cropping to the scaled crop size in Table \ref{tab:auv_data} and resizing the patch to 224x224. The spatial resolutions for the EMR dataset were calculated using the average mission altitude, the calibrated focal length in pixels and the camera sensor pixel size.  The resolutions for the SHR dataset are from \cite{Walker2021Towards}. Note that for the SS09 and SS07 domain pair there was no scaling required as they were from the same AUV platform operating at the same average altitude so have the same spatial resolution already.

\subsection{Training}
The training protocols from previous UDA studies \cite{Zhang2019Domain,Long2017DeepTransferJAN,Zhang2020UMulti} were used.  All labelled source data and unlabelled target data were used for training SymmNet.  The data was augmented by adding a horizontal flip as used in \cite{Zhang2019Domain}.  The same model and training parameters were used for all experiments on both datasets.  

The initial learning rate was 0.02 for the classifier and pooling layer while the pre-trained ResNet50 layers had a learning rate one-tenth lower.  The learning rate followed an annealing strategy as in \cite{Zhang2019Domain} which reduced the learning rate according to \begin{math} \eta_p = \frac{\eta_0}{(1+\alpha p)^\beta}\end{math} where $p$ is the ratio of current epoch to total epochs, $\eta_0$ = 0.02, $\alpha$ = 10 and $\beta$ = 1.5.

Following the two-phase training approach of \cite{Yu2022Efficient} when training the bilinear pooling layer, the learning rate was fixed at 0.1 for the first 3 epochs, followed by a learning rate of 0.02 using the annealing strategy described earlier.  The parameters for the bilinear pooling layer were $a$ and $b$ = 64, $r$ = 16 and $q$ = 4.

The average accuracy from three training runs for classifying all the target data was calculated using the model at the final epoch.  Batch size was 32 and training was for 50 epochs using a Stochastic Gradient Descent optimiser with momentum = 0.9.  

\setlength{\tabcolsep}{3pt}
\settowidth{\rotheadsize}{SymmNet\quad}
\qquad
\renewcommand\theadalign{bc}

\begin{table*}[!ht]
\caption{Classification Accuracy of Target Domain for all Source$\rightarrow$Target pairs with combinations of Scaling, Bilinear Pooling and SymmNet UDA. The best result is shown in red and the second best result is shown in blue.\label{tab:combined_results}}
\centering
   \begin{tabular}{ccc|cccccc|cc}
   \hline
  \multirowcell{-2}[2ex]{\rothead{\\[2.5ex]Scaling}} & \multirowcell{-2}[2ex]{\rothead{\\[2.5ex]Bilinear Pool}} &\multirowcell{-2}[2ex]{\rothead{\\[2.5ex]SymmNet}} & \multicolumn{6}{c|}{\thead{\\[2ex]EMR dataset}} & \multicolumn{2}{c}{\thead{\\[2ex]SHR dataset}}\\[3ex]
   &&&  NG06$\rightarrow$SS07 & SS07$\rightarrow$NG06 & NG06$\rightarrow$SS09 & SS09$\rightarrow$NG06 & SS07$\rightarrow$SS09 & SS09$\rightarrow$SS07 & AE2000f$\rightarrow$Tuna-sand & Tuna-sand$\rightarrow$AE2000f  \\ \hline
        ~ & ~ & ~ & 71.86 & 70.41 & 65.41 & 64.30 & ~ & ~ & 63.92 & 54.52 \\ 
        ~ & \checkmark & ~ & 70.29 & 70.32 & 64.46 & 63.71 & ~ & ~ & 60.79 & 50.88 \\ 
        ~ & ~ & \checkmark & 77.55 & 79.47 & 73.13 & 74.44 & ~ & ~ & 84.44 & 57.44 \\ 
        ~ & \checkmark & \checkmark & 78.88 & 77.22 & 72.97 & \color[rgb]{0,0,1}{74.77} & ~ & ~ & 77.12 & 63.29 \\ \hline
        \checkmark & ~ & ~ & 75.51 & 75.48 & 69.50 & 72.42 & 71.47 & 69.16 & 74.14 & 73.50 \\ 
        \checkmark & \checkmark & ~ & 74.50 & 77.12 & 71.63 & 72.68 & 72.42 & 69.36 & 67.81 & 66.60 \\
        \checkmark & ~ & \checkmark & \color[rgb]{0,0,1}{79.43} & \color[rgb]{1,0,0}{80.19} & \color[rgb]{1,0,0}{76.67} & 74.26 & \color[rgb]{1,0,0}{75.65} & \color[rgb]{0,0,1}{72.64} & \color[rgb]{1,0,0}{93.59} & \color[rgb]{1,0,0}{83.27} \\ 
        \checkmark & \checkmark & \checkmark & \color[rgb]{1,0,0}{80.48} & \color[rgb]{0,0,1}{79.27} & \color[rgb]{0,0,1}{75.26} & \color[rgb]{1,0,0}{76.41} & \color[rgb]{0,0,1}{73.13} & \color[rgb]{1,0,0}{74.50} &  \color[rgb]{0,0,1}{91.91} & \color[rgb]{0,0,1}{78.58} \\ \hline
\end{tabular}
\end{table*}
\setlength{\tabcolsep}{1.4pt} 

\section{Results and Discussion}
\label{sec:result}

\begin{figure*}
    \centering
    \includegraphics[width=0.8\textwidth]{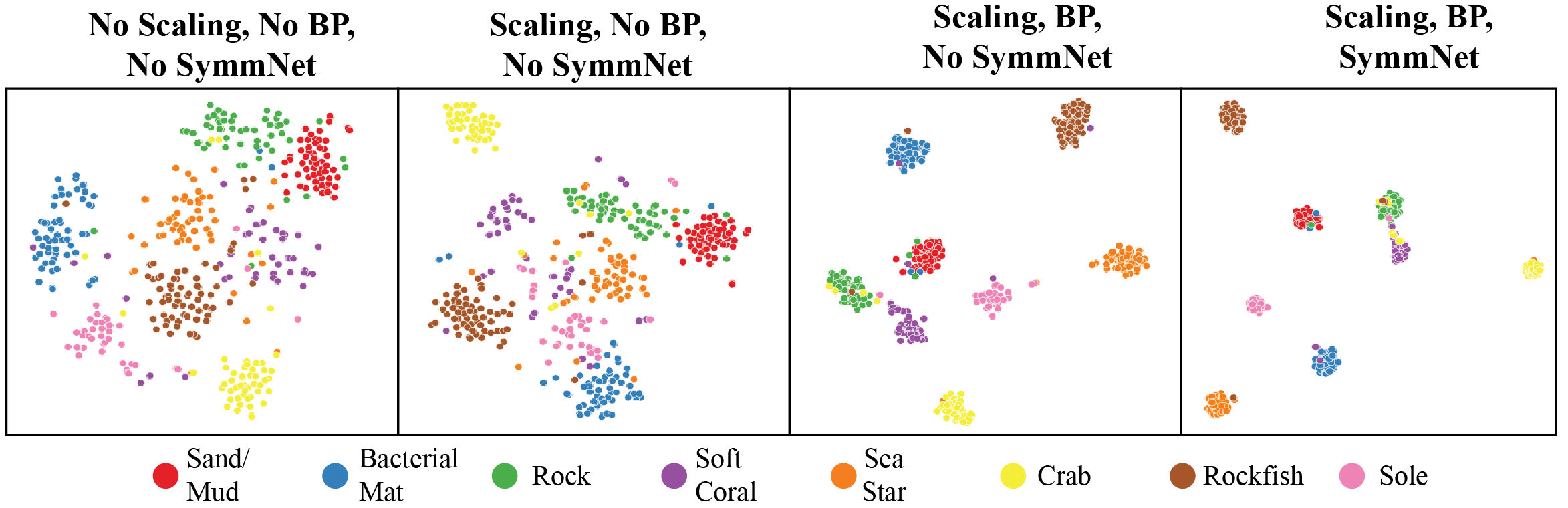}
    \caption{A t-SNE visualization of the features for AUV \emph{Tuna-sand} data for AE2000f$\rightarrow$Tuna-sand source-target pair.  The plots show the effect of adding scaling, bilinear pooling and SymmNet corresponding to lines 1, 5, 6 and 8 in Table \ref{tab:combined_results}.}    
    \label{tsne}
\end{figure*}

Table \ref{tab:combined_results} shows the results of the experiments on each source and target pair.  The average accuracy for the unlabelled target data from the last epoch of training is shown with the best accuracy in red and the second best in blue.  Applying either scaling or SymmNet on their own always produced an increase in accuracy.  Using scaling and SymmNet increased accuracy by up to 28\% compared to using the classifier trained without scaling or domain adaptation. For the majority of source-target domain pairs, the top two results used both scaling and SymmNet while bilinear pooling did not provide a consistent improvement.   

The most significant improvement in accuracy occurred with the SHR dataset where there is a larger difference in spatial resolution between the AUV surveys.  This result demonstrates the ability to use manual annotations from platforms with lower resolution to classify images from higher resolution payloads by reducing the domain shift. This is relevant to real-world, long-term use of survey platforms that are upgraded after extensive annotation of older data sets.

Replacing the ResNet average pooling layer with the bilinear pooling layer only improved accuracy in five of the eight scaled results and none of the unscaled results.  The bilinear pooling layer does not appear to learn a feature that is more discriminative than the average pooling layer that is used by ResNet. This may be due to the hyper-parameters and losses being used to train the network.  A specific training approach and losses for this layer may improve the result.  

Finding the optimal parameters for UDA through model tuning and selection is not possible as there is no labelled target data.  More research on suitable metrics or validation process without labelled target data such as \cite{Ganin2015DomainAdversarial} or \cite{Long2014} may allow the network to be tuned to provide more discriminative features.

Figure~\ref{tsne} shows a t-SNE visualization of the features for the AUV \emph{Tuna-sand} by class for four of the results with the \emph{AE2000f}$\rightarrow$\emph{Tuna-sand} source-target pair.  As can be seen, the clusters for each class become more compact showing features that are more discriminative at a class level as scaling, bilinear pooling and SymmNet are added.  This is particularly evident with the introduction of SymmNet, which is able to use the unlabelled target domain information to find a feature space that is discriminative at a class level despite some of the classes being visually similar.  Some clusters have co-mingled classes showing that the images are not always labelled correctly.  A more accurate source model trained with a larger amount of labelled source data may reduce the number of incorrect labels in these clusters.

\section{Conclusion}
\label{sec:conclusion}
This paper investigated whether performance could be improved for a classifier trained on labelled source data and unlabelled target data from AUV surveys with differing payloads, location and altitudes by applying a framework of resolution scaling, bilinear pooling and SymmNet Unsupervised Domain Adaptation.  Using scaling and SymmNet UDA to classify images from two benthic datasets from different AUV surveys consistently improved accuracy and the separation of classes in the feature space.  The framework can increase the value and longevity of a classifier trained on manual point annotations from previous AUV surveys and applied to higher resolution images from upgraded AUV payloads.  While accuracy was improved with both scaling and SymmNet, by reducing the domain shift between source and target data, tuning of model and training parameters particularly for the bilinear pooling layer may increase these gains.  Other combinations of state-of-the-art UDA such as Source Hypothesis Transfer (SHoT)~\cite{Liang2021SHOTplus} and different implementations of bilinear pooling could be investigated to provide further improvements to the framework. 






\section*{ACKNOWLEDGMENT}
The images from the EMR dataset were part of Australia’s Integrated Marine Observing System (IMOS), enabled by the National Collaborative Research Infrastructure Strategy (NCRIS).  The EMR annotations were based on work undertaken for the Marine Biodiversity Hub, a collaborative partnership supported through funding from the Australian Government’s National Environmental Science Program (NESP).  The images for the SHR dataset were collected during the Schmidt Ocean Institute's FK180731 \#Adaptive Robotics campaign, with support from the Japanese Government's Zipangu in the Ocean Strategic Innovation Program. 


\bibliographystyle{IEEEtran}
\bibliography{bibliography}

\end{document}